\documentclass[10pt,twocolumn,letterpaper]{article}

\usepackage{cvpr}              

\usepackage{algorithm}
\usepackage{algorithmic}
\usepackage{multirow}
\usepackage{bbding}
\usepackage{pifont}
\usepackage{makecell}
\usepackage{color} 
\usepackage{colortbl}
\usepackage[table]{xcolor}
\usepackage{newfloat}
\usepackage{listings}
\usepackage{subcaption} 
\usepackage{relsize}
\definecolor{cvprblue}{rgb}{0.21,0.49,0.74}
\usepackage[pagebackref,breaklinks,colorlinks,allcolors=cvprblue]{hyperref}

\def\paperID{} 
\def\confName{CVPR}
\def\confYear{2026}

\title{Revisiting Token Compression for Accelerating ViT-based Sparse Multi-View \\ 3D Object Detectors}

\author{Mingqian Ji$^{1}$, \quad 
Shanshan Zhang$^{1,2}$\thanks{Corresponding author.}, \quad
Jian Yang$^{1,2}$ \\
\relsize{-0.5} $^1$ PCA Lab, School of Computer Science and Engineering, Nanjing University of Science and Technology \\
\relsize{-0.5} $^2$ PCA Lab, School of Intelligence Science and Technology, Nanjing University \\
\relsize{-1} \{\texttt{mingqianji,shanshan.zhang,csjyang\}@njust.edu.cn}
}

\begin{document}
\maketitle
\begin{abstract}
Vision Transformer (ViT)-based sparse multi-view 3D object detectors have achieved remarkable accuracy but still suffer from high inference latency due to heavy token processing. To accelerate these models, token compression has been widely explored. However, our revisit of existing strategies, such as token pruning, merging, and patch size enlargement, reveals that they often discard informative background cues, disrupt contextual consistency, and lose fine-grained semantics, negatively affecting 3D detection. To overcome these limitations, we propose SEPatch3D, a novel framework that dynamically adjusts patch sizes while preserving critical semantic information within coarse patches. Specifically, we design Spatiotemporal-aware Patch Size Selection (SPSS) that assigns small patches to scenes containing nearby objects to preserve fine details and large patches to background-dominated scenes to reduce computation cost. To further mitigate potential detail loss, Informative Patch Selection (IPS) selects the informative patches for feature refinement, and Cross-Granularity Feature Enhancement (CGFE) injects fine-grained details into selected coarse patches, enriching semantic features. Experiments on the nuScenes and Argoverse 2 validation sets show that SEPatch3D achieves up to \textbf{57\%} faster inference than the StreamPETR baseline and \textbf{20\%} higher efficiency than the state-of-the-art ToC3D-faster, while preserving comparable detection accuracy. Code is available at https://github.com/Mingqj/SEPatch3D.
\end{abstract}
    
\section{Introduction}
\label{sec:intro}
Multi-view 3D object detection has received increasing attention in recent years. This task aims to predict object categories and 3D information from multi-view RGB images, and serves as a critical component in autonomous driving. 

Among various detectors, sparse query-based detectors \cite{detr3d,open,raydn,sparse4d} have recently shown strong performance on public benchmarks. These methods bypass dense BEV construction \cite{bevformer,wu2025see,bevdet4d,wang2025pv,bevdepth} and directly associate learnable object queries with image features, allowing them to focus on object-level information. This design enables sparse detectors to benefit more from high-quality image features, making them particularly well-suited to be paired with Vision Transformers (ViTs) \cite{dosovitskiy2020image}, which offer strong representation capabilities. However, this performance gain comes at the cost of significant inference latency, due to the computational overhead of dense token processing \cite{zeng2024tcformer,zeng2025token,kim2024token,lin2024not}, which has been largely overlooked in previous works. This motivates us to investigate efficient strategies to accelerate ViT-based sparse multi-view 3D detectors.
    \begin{figure}[t]
      \centering
      \begin{subfigure}[t]{0.22\textwidth}
        \centering
        \includegraphics[width=\textwidth]{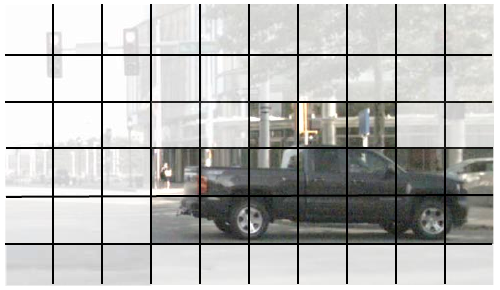}
        \caption{Token pruning}
        \label{token_pruning}
      \end{subfigure}
      \hfill
      \begin{subfigure}[t]{0.22\textwidth}
        \centering
        \includegraphics[width=\textwidth]{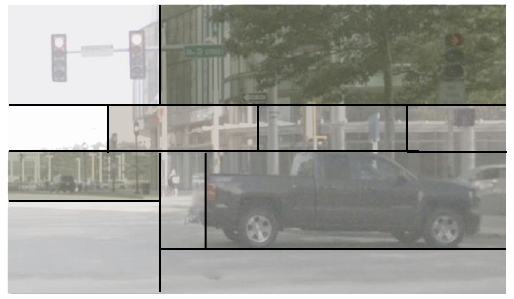}
        \caption{Token merging}
        \label{token_merging}
      \end{subfigure}
      \hfill
      \begin{subfigure}[t]{0.22\textwidth}
        \centering
        \includegraphics[width=\textwidth]{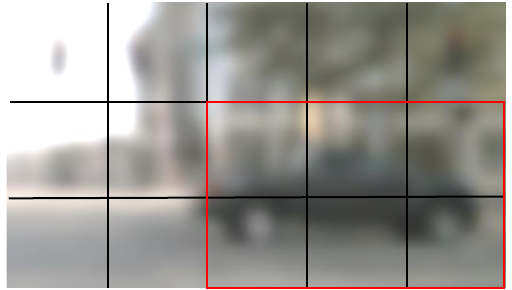}
        \caption{Patch size enlargement}
        \label{patch_enlargement}
      \end{subfigure}
      \hfill
      \begin{subfigure}[t]{0.22\textwidth}
        \centering
        \includegraphics[width=\textwidth]{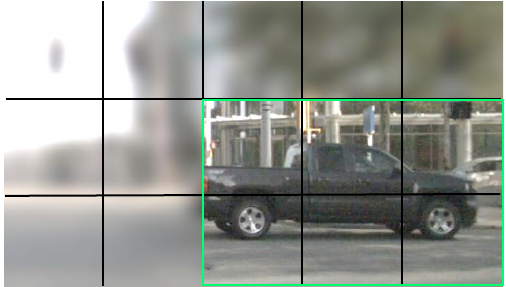}
        \caption{Ours}
        \label{ours}
      \end{subfigure}
      \caption{Visualization of different token compression strategies. (a) Token pruning may remove background cues crucial for hard negatives. (b) Token merging introduces irregular aggregations that disrupt semantic consistency. (c) Enlarging patch size loses fine details (red box), harming accuracy. (d) Our method enhances informative large patches (green box) to retain critical semantics.}
    \end{figure}

\begin{figure}[t!]
        \centering
        \includegraphics[width=8.3cm]{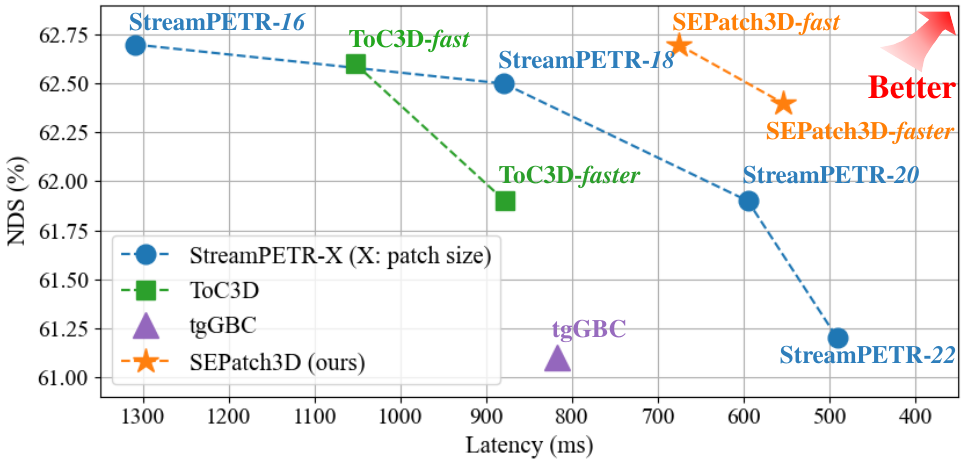}
        \caption{Accuracy-latency under an input resolution of 640 $\times$ 1600 on the nuScenes set. The baseline StreamPETR uses a patch size of 16. As the patch size increases, inference latency consistently decreases, but excessive enlargement leads to accuracy degradation. In contrast, our SEPatch3D-\textit{faster} reduces inference latency by \textbf{57\%} while maintaining competitive detection accuracy.}
        \label{accuracy_latency}
    \end{figure}

Since the computational cost of ViTs largely scales with the number of tokens, token compression has become a straightforward and effective acceleration strategy. Common techniques include token pruning, token merging, and patch size enlargement \cite{zhang2025image}. Among them, token pruning degrades low-importance tokens based on certain heuristics or learned scores. However, this process may lead to the loss of informative background regions (see Fig. \ref{token_pruning}), which are essential for learning hard negative samples in detection tasks. Token merging reduces the token count by combining similar or redundant tokens into more compact representations (see Fig. \ref{token_merging}), but irregular merging disrupts contextual consistency, resulting in semantic information loss. In contrast, patch size enlargement reduces the number of tokens during the patch embedding stage by enlarging the receptive field per token (see Fig. \ref{patch_enlargement}). This strategy preserves the overall semantic contents, making it inherently more compatible with the demands of 3D object detection. However, as shown in Fig. \ref{accuracy_latency}, we observe that simply enlarging the patch size leads to performance degradation when exceeding a certain threshold (e.g., patch size$>$18). This is because coarse-grained patches may overlook fine-grained semantic cues essential for accurate object recognition and localization. The above observations motivate us to design a novel compression strategy, which not only preserves fine-grained semantic information (see Fig. \ref{ours}) crucial for detection, but also is of high efficiency. Building on this insight, our key idea is to reduce the number of patches by enlarging patch sizes while enhancing semantically important regions to mitigate potential information loss.

In this paper, we propose Dynamic Patch Size Selection and Informative Patch Enhancement (\textbf{SEPatch3D}) for efficient ViT-based sparse multi-view 3D object detection. Specifically, we first design the Spatiotemporal-aware Patch Size Selection (SPSS) module, which builds an indicator to estimate the object spatiotemporal distribution. Based on this, we assign small patches to scenes with nearby objects in order to preserve fine-grained details, while large patches are allocated to background-dominated scenes to reduce computational overhead. To further compensate for potential detail loss in coarse patches, we assign entropy-based importance scores to all patches by jointly considering object motions and patch contents, and select the most informative ones via the Informative Patch Selection (IPS) module, and then enhance these selected coarse patches by fusing them with fine-grained patches through a cross-attention via the Cross-Granularity Feature Enhancement (CGFE) module. We evaluate SEPatch3D on the nuScenes \cite{caesar2020nuscenes} and Argoverse 2 \cite{Argoverse2} validation datasets. On nuScenes, our method achieves up to \textbf{57\%} inference speedup over the StreamPETR baseline \cite{wang2023exploring} with less than one point performance degradation. Moreover, SEPatch3D outperforms the state-of-the-art ToC3D-\textit{faster} variant \cite{zhang2024make}, achieving an additional \textbf{20\%} improvement in inference efficiency.

Our contributions are summarized as follows:

\begin{itemize}
    \item We revisit token compression strategies when applied to 3D object detection: token pruning, merging, and patch size enlargement, and analyze their limitations, including the loss of informative backgrounds, disruption of contextual consistency, and degradation of fine-grained semantics. These motivate an efficient compression strategy that better balances accuracy and efficiency.

    \item We propose a novel compression strategy that adaptively enlarges patch sizes according to the spatiotemporal distribution of object queries derived from motion cues. To mitigate detail loss, we introduce an entropy-based patch selection module to identify informative regions, followed by a feature enhancement module that injects fine-grained details into the selected coarse patches.

    \item Experiments on the nuScenes and Argoverse 2 validation sets show that our method achieves faster inference while maintaining competitive accuracy against the baseline and state-of-the-art approaches.
\end{itemize}
\section{Related Work}
\label{sec:formatting}
Since our work focuses on the task of sparse multi-view 3D object detection and token compression for accelerating ViT-based detectors, we first provide a brief overview of token compression techniques for ViTs, followed by recent advances in sparse and efficient multi-view 3D detection.

\subsection{Token Compression for ViTs}
Token compression is a simple and effective strategy to accelerate ViTs, including token pruning, token merging, and patch size enlargement \cite{zhang2025image}.

Token pruning reduces computation by removing low-importance regions or routing them through lightweight processing branches. DynamicViT \cite{rao2021dynamicvit} inserts learnable modules into transformer blocks for progressive pruning. EViT \cite{liang2022not} uses attention weights to route less important tokens into a fused token. Evo-ViT \cite{xu2022evo} adopts a slow-fast update scheme to dynamically retain informative tokens and fuse non-informative tokens. HeatViT \cite{dong2023heatvit} employs a multi-stage training pipeline for token selection. Despite their efficiency, these methods risk discarding background information that provides valuable negative samples for 3D object detection.

Token merging reduces token count by aggregating similar or adjacent tokens. T2T-ViT \cite{yuan2021tokens} performs layer-wise merging through recursive token-to-token transformations. ToMe \cite{bolya2022token} matches and merges semantically similar tokens. PatchMerger \cite{renggli2022learning} employs learned attention weights to merge tokens. MHSA \cite{bian2023multi} identifies crucial tokens based on importance and similarity scores. TcFormer \cite{zeng2022not} introduces a progressive clustering mechanism for layer-by-layer token merging. While these methods are effective, they inevitably disrupt contextual consistency and cause severe information loss.

Patch size enlargement reduces the number of tokens by enlarging patch size to capture a broader receptive field. CvT \cite{wu2021cvt} proposes a hierarchical structure that enlarges patch size at each stage of token embedding. Similarly, SSA \cite{ren2022shunted} gradually increases patch size across attention blocks. TopFormer \cite{zhang2022topformer} builds a multi-scale token pyramid to extract global tokens. In contrast, our approach increases patch size based on the spatiotemporal cues and enhances large coarse patches with finer ones.

\subsection{Sparse Multi-View 3D Object Detection}
Recently, the state-of-the-art methods in multi-view 3D object detection have seen growing interest in sparse query-based methods, which avoid dense BEV construction and instead directly model object-level features through learnable queries.
DETR3D \cite{detr3d} initiates this paradigm by leveraging 3D reference points to sample image features via attention. Sparse4D \cite{sparse4d} enhances temporal modeling with a deformable 4D aggregation module to capture spatiotemporal cues. 3DPPE \cite{shu20233dppe} employs a depth prediction network to generate 3D point positional encodings. SparseBEV \cite{sparsebev} replaces reference points with 3D sparse pillars for improved spatial alignment. StreamPETR \cite{wang2023exploring} introduces a memory queue to propagate temporal information across frames, enabling better tracking and stability. While sparse query-based methods demonstrate impressive performance by directly extracting object-level features from multi-view images, their reliance on powerful ViTs introduces significant inference latency. This motivates us to explore efficient strategies that reduce ViTs latency.

\subsection{Efficient Multi-View 3D Object Detection}
Recently, ToC3D \cite{zhang2024make} pioneered the application of token compression in ViT-based sparse multi-view 3D object detectors. It leverages motion-aware queries to distinguish between foreground and background tokens and performs patch-wise pruning to aggregate information from less important background tokens. However, this approach may discard background regions, limiting the model’s ability to learn from semantically meaningful negative samples. More recently, tgGBC \cite{xu2025accelerate} focuses on accelerating the decoder stage by computing an importance score for each key and pruning keys with lower scores after each decoder layer. While effective to some extent, this method provides limited overall speed-up, as the decoder contributes far less to total inference time than the backbone in ViT-based detectors.

Among existing approaches, ToC3D is the most related to our work. While both aim to improve efficiency through token compression, our framework differs in its design philosophy: instead of relying solely on token importance for pruning, SEPatch3D dynamically adjusts patch granularity and enhances informative regions, enabling better feature preservation and a superior accuracy-efficiency balance.

    \begin{figure*}[t]
        \centering
        \includegraphics[width=1.0\linewidth]{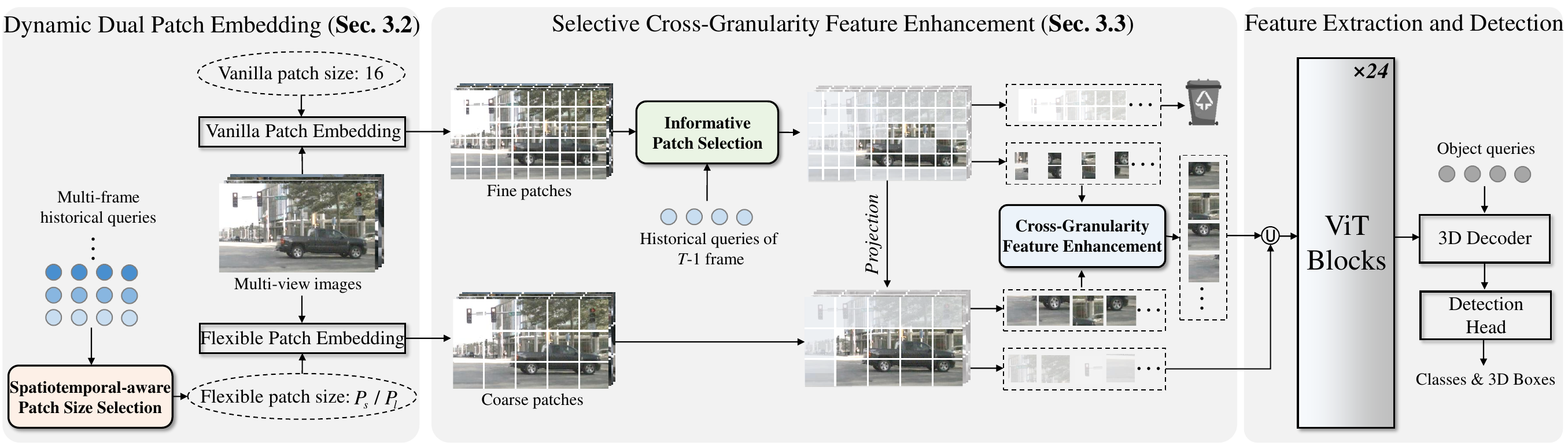}
        \caption{Overview of our method. It consists of two key stages: (1) Dynamic Dual Patch Embedding (Sec. \ref{Dynamic Dual Patch Embedding}), where the SPSS module adaptively adjusts patch sizes according to the spatiotemporal distribution of historical queries; and (2) Selective Cross-Granularity Feature Enhancement (Sec. \ref{Selective Cross-Granularity Feature Enhancement}), where the IPS module selects the most informative patches based on entropy scores, and the CGFE module alleviates information loss by injecting fine-grained informative patches into their corresponding coarse patches.}
        \label{overview}
    \end{figure*}
\section{Methodology}
In this section, we first give an overview of our proposed approach, and then provide a detailed introduction to the main components of our token compression strategy.

\subsection{Overview}
As illustrated in Fig. \ref{overview}, we propose SEPatch3D to accelerate ViT-based sparse multi-view 3D detectors. Below, we briefly describe each stage of the pipeline.

\vspace{-12pt}
\paragraph{Dynamic dual patch embedding.}
Given multi-view images, we first perform vanilla patch embedding by dividing each image into fixed-size patches. In parallel, we dynamically choose a larger patch size for the current frame to perform flexible patch embedding. This patch size is determined by the SPSS module, which dynamically determines patch sizes based on spatiotemporal cues from historical queries, assigning smaller patches to close and growing objects (which present fine details) and larger patches to the far and distant ones (which exhibit few details).

\vspace{-12pt}
\paragraph{Selective cross-granularity feature enhancement.}
To mitigate information loss from coarse patches, the IPS module selects the most informative patches based on entropy scores. Then, these selected coarse patches are refined in the CGFE module using the corresponding fine-grained features, enriching the representation of informative regions.

\vspace{-12pt}
\paragraph{Feature extraction and detection.}
The enhanced and remaining patches are fed to ViT blocks \cite{dosovitskiy2020image} for high-level feature extraction. Then, a set of learnable object queries associates these features across views \cite{petr} via deformable attention. Finally, the aggregated features are passed to a detection head \cite{wang2023exploring} to predict the classes and 3D boxes.

\subsection{Dynamic Dual Patch Embedding} \label{Dynamic Dual Patch Embedding}
\paragraph{Vanilla patch embedding.} We first apply vanilla patch embedding to each multi-view image, dividing each image into fixed-size patches of $16 \times 16$ pixels \cite{fang2024eva}, which are linearly projected to form the initial patch tokens. These small-sized patches will provide fine-grained features for subsequent modules, particularly for enhancing the selected coarse patches in Sec. \ref{CGFE}.

\vspace{-12pt}
\paragraph{Spatiotemporal-aware patch size selection.}
Our findings (see Fig. \ref{accuracy_latency}) show that increasing patch sizes reduces computation, but overly large patches degrade detection performance due to the loss of fine-grained semantic cues. To achieve a better accuracy-efficiency trade-off, we assign smaller patches to scenes with nearby objects to preserve details, and larger patches to distant-dominant scenes to reduce redundant background computation.

To enable this, we employ the depth of object queries as a proxy to guide patch size selection. Specifically, we utilize the object queries from the previous frame $T-1$, denoted as $Q^{T-1} = \{q_1^{T-1}, \cdots, q_M^{T-1}\}$, where $M$ is the number of queries, and compute their 3D positions in the ego-centric coordinate system following StreamPETR \cite{wang2023exploring}. Then, the corresponding depth values $D^{T-1} = \{d_1^{T-1}, \cdots, d_M^{T-1}\}$ are averaged to represent the scene’s spatial layout:
    \begin{equation}
        \bar{D}^{T-1} = \frac{1}{M} \sum_{i=1}^{M} d_i^{T-1}.
    \end{equation}
However, relying solely on spatial cues may lead to abrupt patch size changes across consecutive frames. To ensure temporal stability, we further model the average depth trend across the previous $h$ frames using linear regression \cite{su2012linear}:
    \begin{equation}
        S^{T-1} = \text{Linear}(\bar{D}^{T-h}, \cdots, \bar{D}^{T-1}),
    \end{equation}
where $S^{T-1}$ denotes the slope of the depth trend. Then, we use $\Delta S^{T-1} = S^{T-1} - S^{T-2}$ to capture the variation of the depth trend. Based on both the spatial depth cue $\bar{D}^{T-1}$ and the temporal variation $\Delta S^{T-1}$, the patch size for the current frame $T$ is dynamically determined as:
    \begin{equation}
        P_a^T = 
        \begin{cases}
            P_l, & \text{if } \bar{D}^{T-1} > \theta \text{ and }  \Delta S^{T-1} > 0 \\
            P_s, & \text{if } \bar{D}^{T-1} < \theta \text{ and }  \Delta S^{T-1} < 0 \\
            P_a^{T-1}, & \text{otherwise}
        \end{cases} \quad,
        \label{theta}
    \end{equation}
where $P_s$ and $P_l$ are the pre-defined patch sizes; $\theta$ is a pre-defined depth threshold.
This design ensures an adaptation: (1) When objects are far and becoming smaller ($\bar{D}^{T-1} > \theta \text{ and }  \Delta S^{T-1} > 0 $), we choose the larger patch size to reduce redundant computation of backgrounds; (2) When objects are close and growing ($\bar{D}^{T-1} < \theta \text{ and }  \Delta S^{T-1} < 0$), we choose the relatively smaller patch size for finer detail extraction; (3) Otherwise, we retain the previous patch setting to maintain temporal stability.  


\vspace{-12pt}
\paragraph{Flexible patch embedding.}
After the SPSS module outputs flexible patch sizes, we employ a flexible patch embedding \cite{beyer2023flexivit} to obtain coarse patches.

\subsection{\fontsize{10.2}{12}\selectfont \textbf{Selective Cross-Granularity Feature Enhancement}} \label{Selective Cross-Granularity Feature Enhancement}
\paragraph{Informative patch selection.}
To mitigate the potential loss of critical details caused by coarse patches, we propose an informative patch selection module to identify informative patches for further enhancement. 

\begin{figure}[t]
        \centering
        \includegraphics[width=\linewidth]{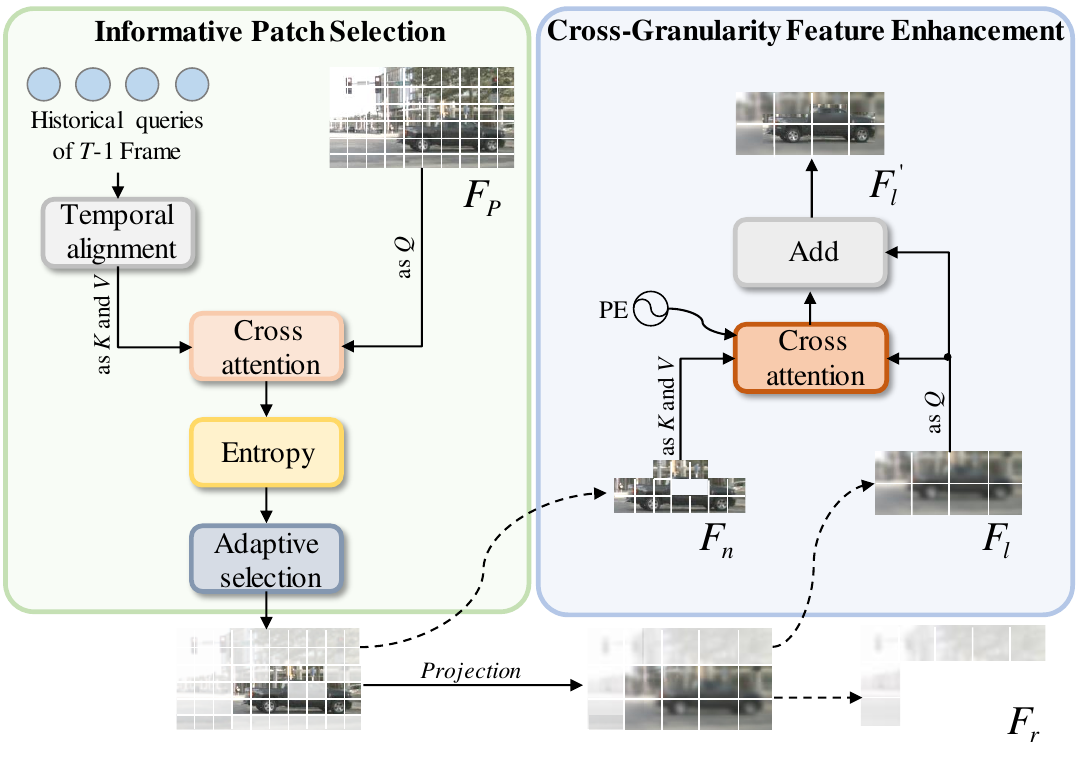}
        \caption{Illustration of the proposed Informative Patch Selection (IPS) and Cross-Granularity Feature Enhancement (CGFE) modules. IPS identifies informative patches based on entropy computed from temporally enhanced features. CGFE refines coarse patches by integrating fine-grained details from fine patches.}
        \label{IPS_CGFE}
    \end{figure}

The pipeline of IPS is shown in Fig. \ref{IPS_CGFE}. Specifically, following StreamPETR \cite{wang2023exploring}, we first align the historical queries with the current frame via motion estimation, obtaining motion-aligned query embeddings $\hat{Q}_T \in \mathbb{R}^{M \times C}$, where $M$ is the number of queries and $C$ is the feature dimension. Given the current patch features $F_p \in \mathbb{R}^{N \times C}$, we enhance each patch feature using a cross-attention:
    \begin{equation}
        F_q = \mathrm{softmax} \left( \frac{F_p\hat{Q_T}^\top}{\sqrt{C}} \right)  \hat{Q_T}.
    \end{equation}
This cross-attention allows the patch features to absorb temporal cues from motion-aligned historical queries, facilitating better foreground object reasoning across frames. 

Since patch features are extracted from shallow layers with limited semantic abstraction, we adopt entropy to identify informative regions, as patches with higher entropy generally correspond to shallow texture-rich or edge areas \cite{guan2022delving,XuHYZ24,abeywickrama2025entrope} that are prone to loss during coarse patch embedding. To achieve this, we first apply $L2$ normalization to patch features across the feature dimension, obtaining $\tilde{F}$. The entropy of $\tilde{F}$ is then computed as:
    \begin{equation}
        \mathcal{H}_j = - \sum_{c=1}^{C} \tilde{F}_{j,c} \log \tilde{F}_{j,c},
    \end{equation}
where $\mathcal{H}_j$ denotes the entropy of the $j$-th patch; $C$ is the feature dimension. Then, instead of selecting the top-$K$ patches \cite{jang2016categorical,zhang2024make}, we adopt an adaptive selection strategy, where patches whose entropy values exceed the mean entropy of all patches in the scene are selected. This avoids the limitations of a fixed top-$K$ approach, allowing the selection to adapt to varying scene complexity.

Notably, the selection is performed on vanilla patches, and the selected indices are projected to the coarse patch space, ensuring spatial alignment across scales. The effectiveness of the design choice is further validated through ablations provided in the supplementary material.

\vspace{-12pt}
\paragraph{Cross-granularity feature enhancement.} \label{CGFE}
After identifying informative regions via the entropy-based patch selection, we introduce a cross-granularity feature enhancement module to improve the quality of coarse patches.

The pipeline of CGFE is shown in Fig. \ref{IPS_CGFE}. Specifically, given the selected vanilla patch features $F_n$ and their corresponding coarse patch features $F_l$, we apply cross-attention to enhance each coarse patch by aggregating detailed information from $F_n$:
    \begin{equation}
        F_e = \mathrm{softmax} \left( \frac{pos(F_l)pos(F_n)^\top}{\sqrt{C}} \right) F_n,
    \end{equation}
where $F_e$ denotes the enhanced coarse patch features; the function $pos(\cdot)$ denotes the addition of the patch positional encoding $\mathrm{PE}$ to the selected patches ($pos(F) = F + \mathrm{PE}$); $C$ is the feature dimension. The coarse patch features act as queries to retrieve fine information from the vanilla patch features, allowing semantic refinement guided by localized details. To preserve the original global structure while injecting fine-level information, we apply a residual connection: $F_l'= F_l + F_e$.
This interaction enables coarse patches to incorporate fine-grained details from vanilla patches, improving feature quality in informative regions. 

\section{Experiments}
This section presents the experimental setups, including the datasets, metrics, and implementation details. We then compare our method with the SOTA methods, followed by ablations, budget-aware analysis, and visualizations.

    \begin{table*}[t]
        \centering
        \caption{Comparisons with the state of the art on the nuScenes validation set. We report the inference time of the whole model as shown in the column ``Infe. Time" and the time of the backbone only as shown in the column ``Back. Time". The default image resolution is 320 $\times$ 800. * and $\dag$ denote image resolutions of 512 × 1408 and 640 × 1600, respectively. $\#$ indicates acceleration in the 3D decoder structure. The difference between our \textit{fast} and \textit{faster} variants lies in the different patch sizes, as detailed in the implementation details.}
        \setlength{\tabcolsep}{1.6mm}{
        \scalebox{0.78}{
        \begin{tabular}{c|c|c c|c c c c c|c c} 
        \hline
        Methods & Backbone & NDS (\%) $\uparrow$ & mAP (\%) $\uparrow$ & mATE $\downarrow$ & mASE $\downarrow$ & mAOE $\downarrow$ & mAVE $\downarrow$ & mAAE $\downarrow$ & Infe. Time (ms) $\downarrow$ & Back. Time (ms) $\downarrow$ \\
        \hline
        DETR3D$\dag$ \cite{detr3d} & ResNet101 & 43.4 & 34.9 & 0.716 & 0.268 & 0.379 & 0.842 & 0.200 & 270.3 & - \\
        3DPPE \cite{shu20233dppe} & VovNet & 44.6 & 39.8 & 0.704 & 0.270 & 0.495 & 0.844 & 0.218 & 125.1 & - \\
        Sparse4D$\dag$ \cite{sparse4d} & ResNet101 & 54.1 & 43.6 & 0.633 & 0.279 & 0.363 & 0.317 & 0.177 & 232.6 & - \\
        SparseBEV* \cite{sparsebev} & ResNet101 & 59.2 & 50.1 & 0.562 & 0.265 & 0.321 & 0.243 & 0.195 & 193.3 & - \\
        StreamPETR \cite{wang2023exploring} & ViT-L & 61.2 & 52.1 & 0.552 & 0.251 & 0.249 & 0.237 & 0.196 & 317.0 & 290.0 \\
        ToC3D-fast \cite{zhang2024make} & ViT-L & 60.9 & 51.7 & 0.552 & 0.250 & 0.268 & 0.229 & 0.195 & 281.0 (-11.4\%) & 253.0 (-12.8\%) \\
        ToC3D-faster \cite{zhang2024make} & ViT-L & 60.5 & 51.3 & 0.562 & 0.250 & 0.265 & 0.230 & 0.203 & 237.2 (-25.2\%) & 209.0 (-28.0\%) \\
        \rowcolor{gray!20}
        \textbf{SEPatch3D-fast} & ViT-L & 61.2 & 52.1 & 0.557 & 0.253 & 0.270 & 0.236 & 0.199 & 250.2 (\textcolor[rgb]{1,0.31,0.51}{-21.1\%}) & 223.6 (\textcolor[rgb]{1,0.31,0.51}{-22.9\%}) \\
        \rowcolor{gray!20}
        \textbf{SEPatch3D-faster} & ViT-L & 60.3 & 51.6 & 0.574 & 0.258 & 0.288 & 0.236 & 0.198 & 194.3 (\textcolor[rgb]{1,0.31,0.51}{-38.7\%}) & 167.9 (\textcolor[rgb]{1,0.31,0.51}{-42.1\%}) \\
        \hline
        StreamPETR$^{\dag}$ \cite{wang2023exploring} & ViT-L & 62.7 & 55.8 & 0.552 & 0.256 & 0.287 & 0.225 & 0.201 & 1309.9 & 1222.4 \\
        ToC3D-fast$^{\dag}$ \cite{zhang2024make} & ViT-L & 62.6 & 54.9 & 0.536 & 0.254 & 0.259 & 0.230 & 0.206 & 1051.9 (-19.7\%) & 964.8 (-21.1\%) \\
        ToC3D-faster$^{\dag}$ \cite{zhang2024make} & ViT-L & 61.9 & 54.3 & 0.560 & 0.257 & 0.230 & 0.234 & 0.201 & 878.5 (-33.0\%) & 791.0 (-35.3\%) \\
        tgGBC$^{\dag}$$^\#$ \cite{xu2025accelerate} & ViT-L & 61.1 & 53.4 & 0.584 & 0.259 & 0.274 & 0.237 & 0.204 & 817.7 (-37.6\%) & - \\
        \rowcolor{gray!20}
        \textbf{SEPatch3D-fast}$^{\dag}$ & ViT-L & 62.7 & 54.5 & 0.526 & 0.256 & 0.235 & 0.236 & 0.207 & 675.4 (\textcolor[rgb]{1,0.31,0.51}{-48.4\%}) & 622.5 (\textcolor[rgb]{1,0.31,0.51}{-49.1\%}) \\
        \rowcolor{gray!20}
        \textbf{SEPatch3D-faster}$^{\dag}$ & ViT-L & 62.4 & 54.2 & 0.523 & 0.255 & 0.252 & 0.242 & 0.195 & 554.4 (\textcolor[rgb]{1,0.31,0.51}{-57.7\%}) & 507.7 (\textcolor[rgb]{1,0.31,0.51}{-58.5\%}) \\
        \hline
        \end{tabular}
            }
        }
        \label{main}
    \end{table*}

    \begin{table*}[t]
        \centering
        \caption{Comparisons with the state of the art on the Argoverse 2 validation set. The default image resolution is 640 $\times$ 960. $\ddag$ indicates that the numbers are from \cite{far3d}. $\diamond$ indicates results reproduced by us.}
        \setlength{\tabcolsep}{3.4mm}{
        \scalebox{0.78}{
        \begin{tabular}{c|c|c c|c c c|c c} 
        \hline
        Methods & Backbone & CDS (\%) $\uparrow$ & mAP (\%) $\uparrow$ & mATE $\downarrow$ & mASE $\downarrow$ & mAOE $\downarrow$ & Infe. Time (ms) $\downarrow$ & Back. Time (ms) $\downarrow$ \\
        \hline
        BEVStereo$^\ddag$ \cite{bevstereo} & VovNet & 10.4 & 14.6 & 0.847 & 0.397 & 0.901 & - & - \\
        SOLOFusion$^\ddag$ \cite{solofusion} & VovNet & 10.6 & 14.9 & 0.934 & 0.425 & 0.779 & - & - \\
        PETR$^\ddag$ \cite{petr} & VovNet & 12.2 & 17.6 & 0.911 & 0.339 & 0.819 & - & - \\
        Sparse4Dv2$^\ddag$ \cite{sparse4dv2} & VovNet & 13.4 & 18.9 & 0.832 & 0.343 & 0.723 & - & - \\
        StreamPETR$^\ddag$ \cite{wang2023exploring} & VovNet & 14.6 & 20.3 & 0.843 & 0.321 & 0.650 & - & - \\ 
        StreamPETR$^\diamond$ \cite{wang2023exploring} & ViT-L & 19.0 & 25.7 & 0.816 & 0.316 & 0.454 & 683.1 & 626.1 \\
        \rowcolor{gray!20}
        \textbf{SEPatch3D-fast} & ViT-L & 18.5 & 25.3 & 0.808 & 0.308 & 0.468 & 483.8 (\textcolor[rgb]{1,0.31,0.51}{-29.2\%}) & 443.1 (\textcolor[rgb]{1,0.31,0.51}{-29.2\%}) \\
        \rowcolor{gray!20}
        \textbf{SEPatch3D-faster} & ViT-L & 18.1 & 24.8 & 0.803 & 0.303 & 0.487 & 405.1 (\textcolor[rgb]{1,0.31,0.51}{-40.7\%}) & 370.7 (\textcolor[rgb]{1,0.31,0.51}{-40.8\%}) \\
        \hline
        \end{tabular}
            }
        }
        \label{argoverse}
    \end{table*}

\subsection{Datasets and Metrics}
We evaluate our method on the nuScenes \cite{caesar2020nuscenes} and Argoverse 2 \cite{Argoverse2} validation sets. For nuScenes, we use the mean Average Precision (mAP) and the nuScenes Detection Score (NDS), where NDS aggregates multiple error types: translation (mATE), scale (mASE), orientation (mAOE), velocity (mAVE), and attribute classification (mAAE). For Argoverse 2, we report mAP and the Composite Detection Score (CDS), which combines mATE, mASE, and mAOE.


\subsection{Implementation Details}
We implement the SEPatch3D with PyTorch \cite{paszke2019pytorch} under the framework MMDetection3D \cite{mmdet3d2020}. We select StreamPETR \cite{wang2023exploring} with a ViT-L backbone \cite{fang2024eva} as our baseline. The $\theta$ in Equ. \ref{theta} is set to 0.6. Following StreamPETR, the number of queries $M$ is set to 64, and the feature dimension $C$ is set to 256. Considering that StreamPETR uses 8 frames of historical information, $h$ is also set to 8. To ensure a fair comparison with StreamPETR in terms of accuracy, we configure two variants of our method: SEPatch3D-\textit{fast} and SEPatch3D-\textit{faster}. On the nuScenes set, under the resolution of $320 \times 800$, the fast variant employs patch sizes $(P_s, P_l)$ of (17, 18), and the faster variant uses (18, 20). Under a higher resolution of $640 \times 1600$, the fast variant is configured with patch sizes of (18, 20), and the faster variant with (20, 22). On the Argoverse 2 set, the resolution is set to $640 \times 960$, and the patch size configuration follows the setting of the higher-resolution experiments on nuScenes. To ensure fairness in runtime, all inference is conducted on a single RTX 3090 GPU. Please refer to the supplementary material for more training details.

\subsection{Main Results}
We compare our method against the baseline StreamPETR and several recent and efficient multi-view 3D object detectors on the nuScenes and Argoverse 2 validation sets.

\vspace{-12pt}
\paragraph{nuScenes.} The main results are shown in Tab. \ref{main}. At an input resolution of 320 $\times$ 800, our SEPatch3D-\textit{fast} achieves 61.2\% NDS and 52.1\% mAP, matching the detection accuracy of StreamPETR while reducing the total inference time by 21.1\% and backbone latency by 22.9\%. To further enhance efficiency, SEPatch3D-\textit{faster} enlarges the patch size, achieving a 38.7\% reduction in runtime with only a 0.9\% drop in NDS, and exhibits a similar runtime cost to SparseBEV \cite{sparsebev} with a ResNet101 \cite{he2016deep} backbone. Moreover, compared with the state-of-the-art ToC3D-\textit{faster}, our SEPatch3D-\textit{faster} provides comparable detection accuracy while being 17.6\% faster in total inference time, demonstrating a more favorable trade-off between efficiency and accuracy. At a higher resolution of 640 $\times$ 1600, SEPatch3D-\textit{fast} achieves 62.7\% NDS and 54.5\% mAP, matching StreamPETR, while significantly reducing the total inference time by 48.4\% and backbone computation by 49.1\%. Similarly, SEPatch3D-\textit{faster} further reduces latency by 57.7\% overall and 58.5\% on the backbone, while maintaining competitive accuracy. Compared with other fast variants like ToC3D-\textit{faster} and tgGBC \cite{xu2025accelerate}, SEPatch3D-\textit{faster} achieves faster speedup while maintaining similar detection quality. These results confirm that our SEPatch3D reduces computational runtime while preserving detection accuracy across various input resolutions.

\vspace{-12pt}
\paragraph{Argoverse 2.} As shown in Tab. \ref{argoverse}, our \textit{fast} variant achieves 18.5\% CDS and 25.3\% mAP, closely matching the accuracy of the baseline StreamPETR while reducing the total inference time by 29.2\%. The \textit{faster} variant further enlarges the patch size, leading to a 40.7\% reduction in total runtime with less than a one-point drop in CDS. Compared with previous VovNet \cite{lee2019energy}-based detectors, our ViT-based SEPatch3D achieves substantially higher detection accuracy while offering significant speedup. These results demonstrate that our method generalizes well across different datasets and resolutions, consistently providing a superior trade-off between accuracy and inference efficiency.

    \begin{table}[t]
        \centering
        \caption{Generalization across sparse multi-view 3D detectors.}
        \setlength{\tabcolsep}{0.7mm}{
        \scalebox{0.79}{
        \begin{tabular}{c|c|c c|c} 
        \hline
        Methods & Backbone & NDS (\%) $\uparrow$ & mAP (\%) $\uparrow$ & Infe. Time (ms) $\downarrow$ \\
        \hline
        StreamPETR \cite{wang2023exploring} & ViT-L & 61.2 & 52.1 & 317.0 \\
        \rowcolor{gray!20}
        Ours-fast & ViT-L & 61.2 & 52.1 & 250.2 (\textcolor[rgb]{1,0.31,0.51}{-21.1\%}) \\
        \rowcolor{gray!20}
        Ours-faster & ViT-L & 60.3 & 51.6 & 194.3 (\textcolor[rgb]{1,0.31,0.51}{-38.7\%}) \\
        \hline
        DETR3D \cite{detr3d} & ViT-L & 57.4 & 48.9 & 344.9 \\
        \rowcolor{gray!20}
        Ours-fast & ViT-L & 57.4 & 48.8 & 268.8 (\textcolor[rgb]{1,0.31,0.51}{-22.1\%}) \\
        \rowcolor{gray!20}
        Ours-faster & ViT-L & 57.0 & 48.4 & 216.2 (\textcolor[rgb]{1,0.31,0.51}{-37.3\%}) \\
        \hline
        Sparse4Dv2 \cite{sparse4dv2} & ViT-L & 59.4 & 50.4 & 325.8 \\
        \rowcolor{gray!20}
        Ours-fast & ViT-L & 59.3 & 50.2 & 261.5 (\textcolor[rgb]{1,0.31,0.51}{-19.7\%}) \\
        \rowcolor{gray!20}
        Ours-faster & ViT-L & 58.7 & 50.0 & 204.4 (\textcolor[rgb]{1,0.31,0.51}{-37.3\%}) \\
        \hline
        \end{tabular}
            }
        }
        \label{sparse_detectors}
    \end{table}

    \begin{table}[t]
        \centering
        \caption{Ablations of each component.}
        \setlength{\tabcolsep}{0.8mm}{
        \scalebox{0.82}{
        \begin{tabular}{c|c c|c|c} 
        \hline
        Methods & NDS (\%) $\uparrow$ & mAP (\%) $\uparrow$ & Infe. Time (ms) $\downarrow$ & Parameters (M) \\
        \hline
        Baseline & 61.2 & 52.1 & 317.0 & 316.62 \\
        + SPSS & 58.8 & 50.8 & 189.1 (-40.3\%) & 318.77 (+0.6\%) \\
        + CGFE & 60.4 & 51.7 & 209.6 (-33.9\%) & 325.58 (+2.8\%) \\
        + IPS & 60.3 & 51.6 & 194.3 (-38.7\%) & 328.13 (+3.6\%) \\
        \hline
        \end{tabular}
            }
        }
        \label{each_component}
    \end{table}

\subsection{Ablation Studies}
In this subsection, we conduct ablations to evaluate the generalizaiton and effectiveness of SEPatch3D. The default variants are built upon SEPatch3D-\textit{faster} with an image resolution of 320 $\times$ 800 on the nuScenes validation set. The inference is performed on a single RTX 3090 GPU.


\vspace{-14pt}
\paragraph{Generalization to different sparse multi-view 3D object detectors.}
As shown in Tab. \ref{sparse_detectors}, we evaluate SEPatch3D on multiple sparse multi-view 3D detectors, including StreamPETR \cite{wang2023exploring}, DETR3D \cite{detr3d}, and Sparse4Dv2 \cite{sparse4dv2}. Across all detectors, our SEPatch3D-\textit{fast} and \textit{faster} consistently reduce inference latency while maintaining competitive accuracy. When applied to DETR3D, SEPatch3D-\textit{fast} and \textit{faster} reduce latency by 22.1\% and 37.3\%, respectively, with negligible performance drop. Similar trends are observed for Sparse4Dv2, where our methods achieve up to 37.3\% speedup while preserving detection quality. These results demonstrate that our SEPatch3D generalizes well across different sparse detection frameworks, confirming that our improvements are broadly compatible rather than tightly coupled to a specific detector design.

\vspace{-12pt}
\paragraph{Generalization to different ViT-based image encoders.}
As shown in Tab.\ref{image_encoders}, we evaluate the generalization of our method across different ViT-based image encoders. With the lightweight ViT-B backbone pretrained on SAM \cite{kirillov2023segment}, our SEPatch3D-\textit{fast} and \textit{faster} achieve comparable accuracy to StreamPETR while reducing inference latency by 19.6\% and 35.0\%, respectively. We further test on ToC3DViT-L, which is trained via fine-tuning on ToC3D \cite{zhang2024make}. In this setting, our SEPatch3D-\textit{fast} and \textit{faster} still deliver consistent acceleration, lowering latency by 23.8\% and 37.8\%. These results confirm that our efficiency gains are not tied to a specific backbone but generalize well across diverse ViT-based encoders. We attribute this generalization to our improvements at the patch embedding level, which ensure compatibility with various feature extraction blocks.

\vspace{-12pt}
\paragraph{Impact of each component.} As shown in Tab. \ref{each_component}, introducing the SPSS module reduces inference time by 40.3\% with only a 0.6\% increase in parameters, but performance drops to 58.8\% NDS and 50.8\% mAP, indicating that coarse partitioning without refinement may discard useful semantics. Adding the CGFE module effectively restores accuracy to 60.4\% NDS and 51.7\% mAP while still delivering a 33.9\% speed-up, with overall parameters increasing by 2.8\% compared to the baseline, which is acceptable given its role in enhancing informative regions. Finally, the IPS module further improves the trade-off, achieving 60.3\% NDS and 51.6\% mAP with a 38.7\% runtime reduction, while the total parameters rise by only 3.6\%. Overall, SPSS and IPS mainly contribute to efficiency, whereas CGFE is essential for accuracy, and all modules together introduce only modest parameter growth over the baseline.

    \begin{table}[t]
        \centering
        \caption{Generalization across ViT-based image encoders.}
        \setlength{\tabcolsep}{0.35mm}{
        \scalebox{0.79}{
        \begin{tabular}{c|c|c c|c} 
        \hline
        Methods & Backbone & NDS (\%) $\uparrow$ & mAP (\%) $\uparrow$ & Infe. Time (ms) $\downarrow$  \\
        \hline
        StreamPETR \cite{wang2023exploring} & ViT-B & 54.4 & 43.8 & 125.5 \\
        \rowcolor{gray!20}
        Ours-fast & ViT-B & 54.1 & 43.5 & 100.9 (\textcolor[rgb]{1,0.31,0.51}{-19.6\%}) \\
        \rowcolor{gray!20}
        Ours-faster & ViT-B & 53.2 & 43.0 & 81.6 (\textcolor[rgb]{1,0.31,0.51}{-35.0\%}) \\
        \hline
        StreamPETR \cite{wang2023exploring} & ViT-L & 61.2 & 52.1 & 317.0 \\
        \rowcolor{gray!20}
        Ours-fast & ViT-L & 61.2 & 52.1 & 250.2 (\textcolor[rgb]{1,0.31,0.51}{-21.1\%}) \\
        \rowcolor{gray!20}
        Ours-faster & ViT-L & 60.3 & 51.6 & 194.3 (\textcolor[rgb]{1,0.31,0.51}{-38.7\%}) \\
        \hline
        StreamPETR \cite{wang2023exploring} & ToC3DViT-L & 60.9 & 51.7 & 281.0 \\ 
        \rowcolor{gray!20}
        Ours-fast & ToC3DViT-L & 60.3 & 51.3 & 213.9 (\textcolor[rgb]{1,0.31,0.51}{-23.8\%}) \\
        \rowcolor{gray!20}
        Ours-faster & ToC3DViT-L & 59.9 & 50.9 & 174.8 (\textcolor[rgb]{1,0.31,0.51}{-37.8\%}) \\
        \hline
        \end{tabular}
            }
        }
        \label{image_encoders}
    \end{table} 
    \begin{table}[t]
        \centering
        \caption{Ablations of patch size selection in the SPSS module.}
        \setlength{\tabcolsep}{0.6mm}{
        \scalebox{0.82}{
        \begin{tabular}{c|c|c c|c} 
        \hline
        Patch Size & Patch Numbers & NDS (\%) $\uparrow$ & mAP (\%) $\uparrow$ & Infe. Time (ms) $\downarrow$ \\
        \hline
        18 & 810 & 60.4 & 51.8 & 242.6 \\
        20 & 640 & 59.3 & 50.3 & 167.7 \\
        (18, 20) & (810, 640) & 60.3 & 51.6 & 194.3 \\
        \hline
        \end{tabular}
            }
        }
        \label{SPSS}
    \end{table}

\vspace{-12pt}
\paragraph{Impact of adaptive patch size selection.} 
To further evaluate the effectiveness of adaptive patch size selection in the SPSS module, we compare it against fixed patch sizes. As shown in Tab. \ref{SPSS}, using a fixed patch size of 18 with 810 patches yields the highest accuracy of 60.4\% NDS and 51.8\% mAP, but incurs a high latency of 242.6 ms. Increasing the patch size to 20 reduces the number of patches to 640 and latency to 167.7 ms, at the expense of decreased accuracy. In contrast, our patch size selection strategy dynamically adjusts patch sizes between 18 and 20, achieving competitive performance of 60.3\% NDS and 51.6\% mAP while reducing latency to 194.3 ms. These results demonstrate that adaptive patch size selection offers an effective trade-off between accuracy and efficiency.

\begin{figure}[t]
        \centering
        \captionsetup{skip=2pt}
        \includegraphics[width=0.95\linewidth]{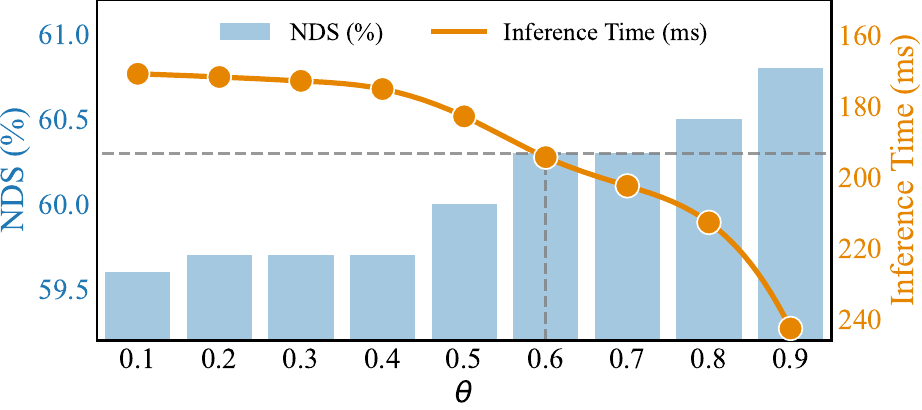}
        \caption{Effect of the threshold $\theta$ in Equ. \ref{theta} on NDS and runtime.}
        \label{theta_nds_inference}
    \end{figure}

\vspace{-12pt}
\paragraph{Impact of depth distribution threshold $\theta$ in Equ. \ref{theta}.}
Fig. \ref{theta_nds_inference} illustrates how the threshold $\theta$ affects the accuracy and efficiency. The blue bars indicate NDS (\%) for each $\theta$, while the orange curve shows inference time (ms). Smaller $\theta$ values assign larger patches to more scenes, reducing inference time but slightly lowering accuracy. Larger $\theta$ values emphasize smaller patches, improving NDS at the cost of increased latency. We select this parameter, $\theta=0.6$, as a balanced setting that achieves the best trade-off between accuracy and inference time. Notably, $\theta=0.6$ remains unchanged on the Argoverse 2 dataset, demonstrating its general applicability across different datasets.

\vspace{-12pt}
\paragraph{Impact of entropy-based patch selection.}
As shown in Tab. \ref{patch_selection}, prior score- \cite{rao2021dynamicvit} and attention-based \cite{wang2023exploring} token selection strategies reduce inference time to around 200\,ms but degrade performance. In contrast, our entropy-based selection achieves the better performance of 60.3\% NDS and 51.6\% mAP. We analyse that our method performs selection at shallow feature stages, where learned scores or attention weights become unreliable due to insufficient high-level semantics. In contrast, entropy offers a measure of information content for shallow features, enabling more reliable identification of informative regions. Therefore, this allows our approach to enhance more meaningful patches, yielding a superior efficiency-accuracy trade-off.

    \begin{table}[t]
        \centering
        \caption{Ablations of patch selection basis in the IPS module.}
        \setlength{\tabcolsep}{2.6mm}{
        \scalebox{0.82}{
        \begin{tabular}{c|c c|c} 
        \hline
        Methods & NDS (\%) $\uparrow$ & mAP (\%) $\uparrow$ & Infe. Time (ms) $\downarrow$  \\
        \hline
        - (Baseline) & 61.2 & 52.1 & 317.0 \\
        Score \cite{rao2021dynamicvit} & 59.8 & 51.2 & 194.3 \\
        Attention \cite{wang2023exploring} & 59.6 & 51.5 & 196.9 \\
        Entropy (ours) & 60.3 & 51.6 & 194.3 \\
        \hline
        \end{tabular}
            }
        }
        \label{patch_selection}
    \end{table}

\subsection{Budget-Aware Analysis}
\begin{figure}[t]
        \centering
        \captionsetup{skip=2pt}
        \includegraphics[width=1.0\linewidth]{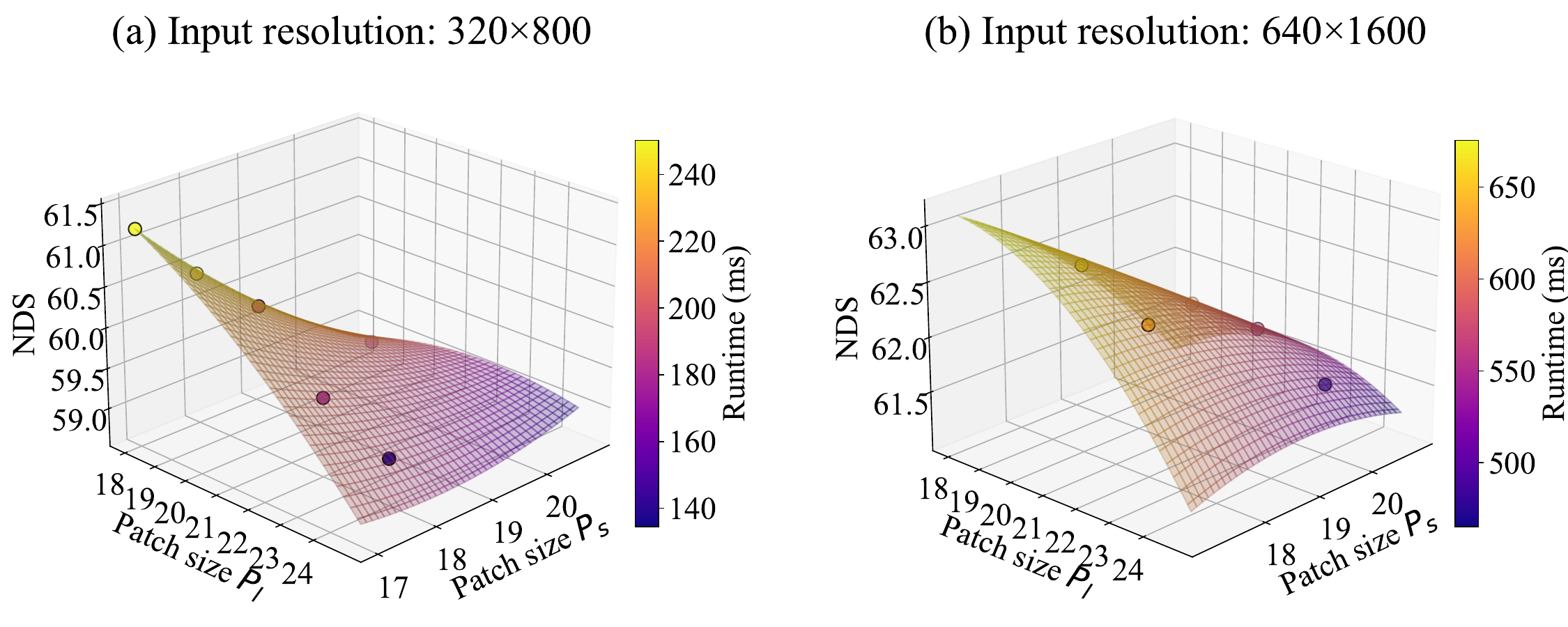}
        \caption{Budget-aware analysis of accuracy-latency under the input resolutions of $320 \times 800$ and $640 \times 1600$ on nuScenes. Colors denote runtime while the surface height indicates NDS.
        }
        \label{budget_analysis}
    \end{figure}

In practical deployment, traversing all patch configurations is time-consuming. To address this, we conduct a budget-aware analysis that predicts patch sizes under target accuracy and runtime. We fit both NDS and runtime as functions of patch sizes $(P_s, P_l)$ with a polynomial regression model (chosen because runtime varies roughly quadratically with fixed patch sizes). Formally, the fitting functions are $f_{\text{NDS}}(P_s, P_l) = \sum_{i=0}^{d}\sum_{j=0}^{d-i} a_{ij} P_s^i P_l^j$ and $f_{\text{Time}}(P_s, P_l) = \sum_{i=0}^{d}\sum_{j=0}^{d-i} b_{ij} P_s^i P_l^j$, where $d$ is the polynomial degree, and $a_{ij}$ and $b_{ij}$ are learned coefficients. The optimal configuration under budgets is obtained by solving $(P_s,P_l)=\arg\min_{(P_s,P_l)} [(f_{\text{Time}}(P_s,P_l)-B_{\text{Time}})^2+(f_{\text{NDS}}(P_s,P_l)-B_{\text{NDS}})^2]$, where $B_{\text{Time}},B_{\text{NDS}}$ denote the runtime and accuracy budgets. As shown in Fig. \ref{budget_analysis}, the fitted algebraic surface supports interpolation to unseen configurations and direct search under constraints. This demonstrates that our analysis provides a practical tool to guide patch selection, balancing efficiency and accuracy without training on all possible configurations.


    \begin{figure}[t]
        \centering
        \captionsetup{skip=2pt}
        \includegraphics[width=1.0\linewidth]{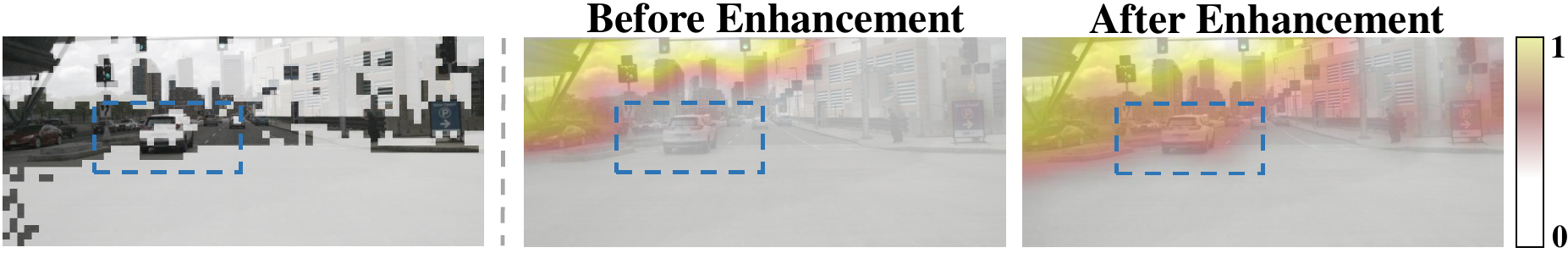}
        \caption{Visualization of informative regions and feature responses. The blue boxes indicate regions with notable differences.}
        \label{vis}
    \end{figure}

\subsection{Visualization Analysis}
As shown in Fig. \ref{vis}, we visualize the informative patches. We can observe that our IPS module effectively highlights regions rich in informative foreground content, such as vehicles and pedestrians, while filtering out redundant areas, such as roads and flat object interiors, demonstrating its capability in selecting truly informative regions. Moreover, we visualize the feature responses. We can observe that the feature activations of the informative patches are significantly enhanced after introducing the CGFE module, further validating the effectiveness of this module. More visualization results are provided in the supplemental material.


\section{Conclusion}
In this paper, we propose SEPatch3D, a novel framework for accelerating ViT-based sparse multi-view 3D object detectors. Specifically, by adaptively enlarging patch sizes, selecting informative regions, and enhancing the selected coarse features SEPatch3D achieves a favorable balance between accuracy and efficiency. Experiments on the nuScenes and Argoverse 2 validation sets demonstrate the effectiveness of our approach. Finally, we hope SEPatch3D will further advance research on efficient token compression for ViT-based sparse multi-view 3D object detectors.

\section*{Acknowledgment}
This research is supported by the National Natural Science Foundation of China (Grant No. 62322602), and the Natural Science Foundation of Jiangsu Province, China (Grant No. BK20230033).

\clearpage
\setcounter{page}{1}
\maketitlesupplementary
In this document, we first provide more implementation details (Sec. \ref{More implementation details}). Then, we present additional experiments (Sec. \ref{more experiments}) and more visualization results (Sec. \ref{More visualization results}) to further validate the effectiveness of our method. Finally, we discuss the limitations of our approach and provide insights into potential future research directions (Sec. \ref{limitation}).

\section{More Implementation Details} \label{More implementation details}
During training, on the nuScenes set, we adopt the default settings of StreamPETR \cite{wang2023exploring}, including the optimizer, learning rate, and data augmentation strategies. The model is trained for a total of 36 epochs on 4 $\times$ RTX 3090 GPUs. Specifically, the first 24 epochs are used to train the detector with the flexible patch embedding module \cite{beyer2023flexivit}, aiming to ensure the detector achieves normal performance under a single patch size. The subsequent 12 epochs focus on the training of our proposed three modules. On the Argoverse 2 set, the model is trained for a total of 12 epochs on 4 $\times$ RTX 3090 GPUs. The optimizer, learning rate, and data augmentation strategies follow the configurations in Far3D \cite{far3d}. 

Since historical queries are required by the SPSS and IPS modules, SPSS adopts a predefined initial patch size ($P_s$) at the first frame as a safe fallback to ensure stable performance, and is activated in subsequent frames once temporal information becomes available. Meanwhile, IPS relies solely on current-frame image features to guide informative region selection in the absence of historical cues.

\section{Additional Experiments} \label{more experiments}
\paragraph{Comparison to 2D token compression methods.}
We further compare SEPatch3D with representative 2D token compression approaches, including SparseDETR \cite{roh2021sparse}, Cropr \cite{bergner2025token}, and SViT \cite{liu2024revisiting}, which perform token pruning at 2D tasks (e.g., classification and object detection). For fair comparison, we re-implement these methods on StreamPETR and apply token pruning at layers [6, 12, 18] with a fixed pruning ratio of 0.5, resulting in similar inference time. As shown in Tab. \ref{token_pruning}, although token pruning effectively reduces computation, it consistently leads to a noticeable performance degradation.
In contrast, our method achieves up to +2 \textit{pp} NDS and +1 \textit{pp} mAP under similar latency. This highlights the advantage of patch-level enlargement over token-level pruning and quantitatively validates the core motivation of our design.

    \begin{table}[h]
        \centering
        \caption{\scalebox{1}{Comparison to token pruning methods.}}
        \setlength{\tabcolsep}{1.8mm}{
        \scalebox{0.78}{
        \begin{tabular}{c |c c|c} 
        \hline
        Methods & NDS (\%) $\uparrow$ & mAP (\%) $\uparrow$ & Inference Time (ms) $\downarrow$ \\
        \hline
        StreamPETR & 61.2 & 52.1 & 317.0 \\
        \hline
        + SparseDETR \cite{roh2021sparse} & 59.5 & 51.2 & 251.3 \\
        + Cropr \cite{bergner2025token} & 60.1 & 51.4 & 251.6 \\
        + SViT \cite{liu2024revisiting} & 60.3 & 51.1 & 250.0 \\
        \hline
        \rowcolor{gray!20}
        Ours-fast & 61.2 & 52.1 & 250.2 \\
        \hline
        \end{tabular}
            }
        }
        \label{token_pruning}
    \end{table}

\begin{figure}[h]
    \centering
    \begin{minipage}[h]{0.38\linewidth}
        \centering
        \captionsetup{labelformat=empty}
        \captionof{table}{\scalebox{0.8}{\parbox{1.25\textwidth}{Table 2. Comparison w.r.t. mFP @ recall=0.2.}}}
        \setlength{\tabcolsep}{1.8mm}{
        \scalebox{0.6}{
        \begin{tabular}{c|c|c} 
        \hline
        Methods & Ratio & mFP (\%) $\downarrow$ \\
        \hline
        StreamPETR & 0 & 35.4 \\
        \hline
        + Cropr & 0.5 & 53.2 \\
        \hline
        \rowcolor{gray!20}
        Ours-fast & 0 & 40.5 \\
        \hline
        \end{tabular}
            }
        }
        \label{corpr_ratio_tab}
    \end{minipage}
    \hfill
    \begin{minipage}[h]{0.59\linewidth}
        \centering
        \includegraphics[width=\linewidth]{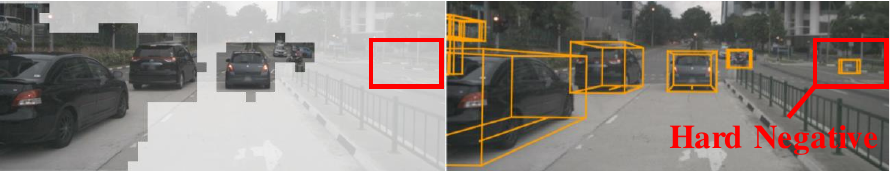}
        \captionsetup{labelformat=empty}
        \captionof{figure}{\scalebox{0.8}{\parbox{1.25\textwidth}{Figure 1. Background patches contain useful hard negative information for 3D detection.}}}
        \label{corpr_ratio_fig}
    \end{minipage}
    \end{figure}

\vspace{-12pt}
\paragraph{Quantitative validation of hard negatives in token pruning.}
To provide a quantitative analysis of how token pruning affects false positive predictions in sparse multi-view 3D detection, we compare StreamPETR equipped with Cropr \cite{bergner2025token} and our SEPatch3D framework in terms of mFP under a fixed recall level. As shown in Tab. \ref{corpr_ratio_tab}, applying token pruning with Cropr (pruning ratio $0.5$) leads to a substantial increase in mFP, from $35.4\%$ to $53.2\%$, indicating that aggressive removal of background tokens significantly degrades the detector’s ability to suppress hard negatives.
In contrast, SEPatch3D maintains a much lower mFP ($40.5\%$) while achieving comparable inference efficiency, demonstrating improved robustness to background-induced false alarms. We further visualize the effect in Fig. \ref{corpr_ratio_fig}, where many background patches pruned by Cropr correspond to regions that provide useful negative evidence. Although these patches do not directly contribute to foreground object representations, they play an important role in constraining the decision boundary and suppressing hard negatives in sparse query-based 3D detectors. These results quantitatively support our design choice of retaining all background patches and avoiding token-level pruning.

\begin{figure}[ht]
        \centering
        \captionsetup{skip=2pt}
        \includegraphics[width=0.95\linewidth]{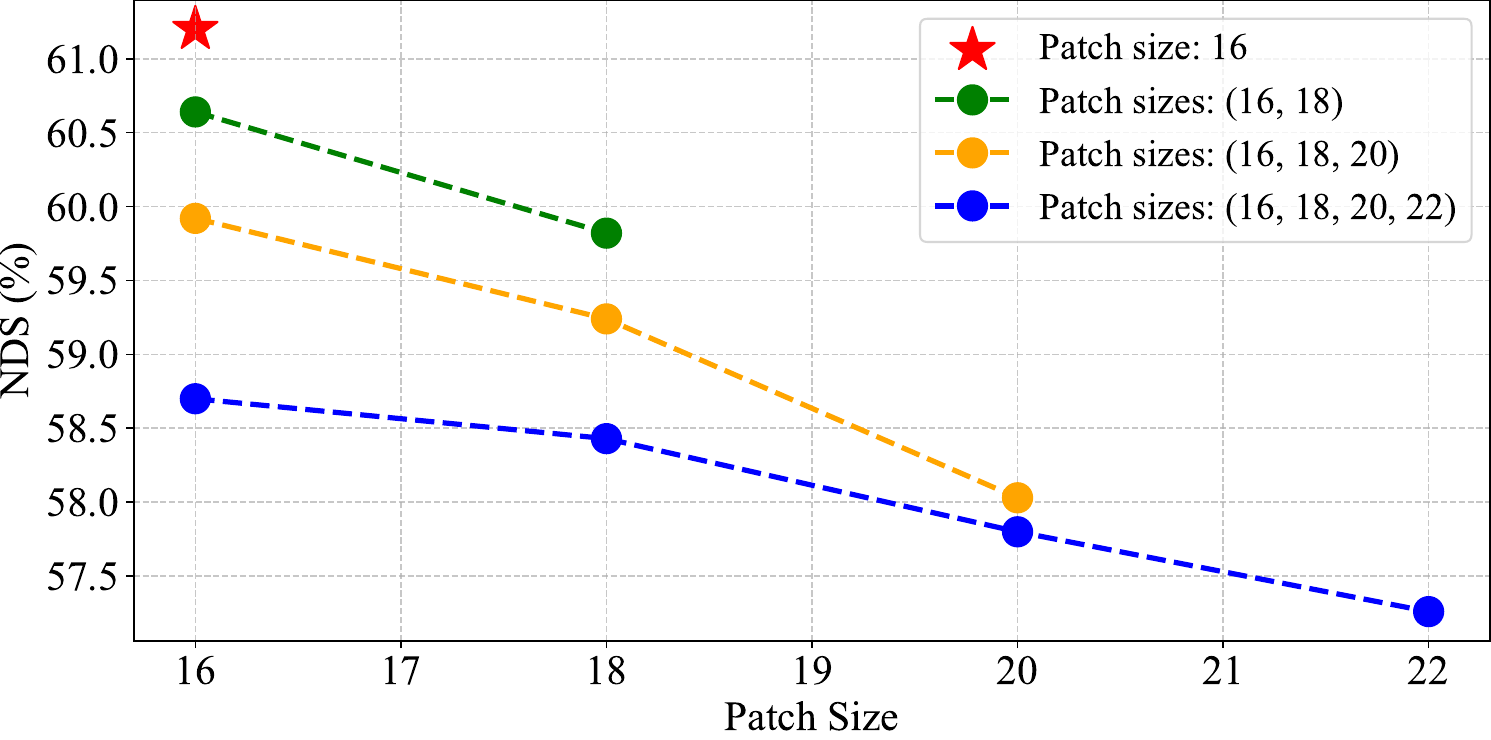}
        \caption{Impact of patch size diversity on detection performance.}
        \label{flexivit}
    \end{figure}

\vspace{-12pt}
\paragraph{Impact of patch size diversity on detection performance in response to: ``why do we only use two patch sizes instead of more?".} As shown in Fig. \ref{flexivit}, using two patch sizes (16, 18) achieves comparable accuracy to the single-size baseline.  However, increasing the number of selectable patch sizes to three (16, 18, 20) or four (16, 18, 20, 22) leads to a gradual decrease in performance measured at patch size 16, with the degradation becoming more pronounced as patch size diversity increases. We analyze that introducing too many selectable patch sizes increases the complexity of token distributions, which forces the detection head to generalize across multiple granularities but weakens optimization for any specific one. In this case, the detection head struggles to learn specific features well for individual patch configurations, leading to suboptimal performance. Therefore, adopting fewer of the two patch sizes not only allows effective dynamic adjustment of patches but also preserves strong specialization, making it an optimal design choice for our SEPatch3D.

\vspace{-12pt}
\paragraph{Impact of spatiotemporal cues in the SPSS module.}
In Tab. \ref{ablation-SPSS}, we analyze the contribution of spatial and temporal cues in the SPSS module. When either spatial or temporal cues are used individually, the inference time is notably reduced compared to the baseline, demonstrating that adaptive patch size selection effectively improves efficiency. However, both configurations suffer from performance degradation in NDS and mAP, indicating that relying on a single cue is insufficient for accurate object motion estimation. When both cues are jointly incorporated, the SPSS module achieves a good trade-off between accuracy and efficiency, further validating the effectiveness of the SPSS module.

    \begin{table}[ht]
        \centering
        \caption{Ablations of spatiotemporal cues in the SPSS module.}
        \setlength{\tabcolsep}{1.1mm}{
        \scalebox{0.82}{
        \begin{tabular}{c c|c c|c} 
        \hline
        Spat. Cues & Temp. Cues & NDS (\%) $\uparrow$ & mAP (\%) $\uparrow$ & Infe. Time (ms) $\downarrow$  \\
        \hline
        & & 61.2 & 52.1 & 317.0 \\
        \ding{51} & & 59.7 & 51.2 & 195.5 \\
        & \ding{51} & 59.5 & 51.3 & 191.3 \\
        \ding{51} & \ding{51} & 60.3 & 51.6 & 194.3 \\
        \hline
        \end{tabular}
            }
        }
        \label{ablation-SPSS}
    \end{table}

\begin{figure}[h]
    \centering
    \begin{minipage}[h]{0.46\linewidth}
        \centering
        \includegraphics[width=\linewidth]{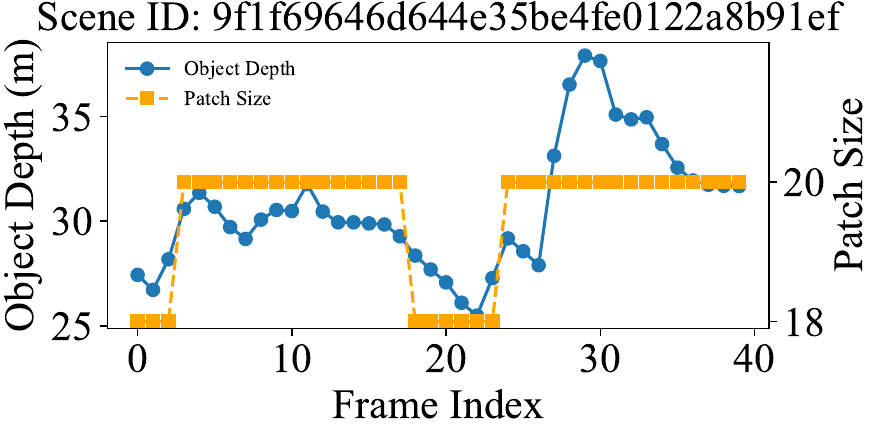}
        \captionsetup{skip=2pt}
        \vspace{-15pt}
        \captionsetup{labelformat=empty}
        \caption{\scalebox{0.8}{Figure 3. Temporal distribution.}}
        \label{DistributionFig}
    \end{minipage}
    \hfill
    \begin{minipage}[h]{0.5\linewidth}
        \centering
        \captionsetup{labelformat=empty}
        \captionof{table}{\scalebox{0.8}{\parbox{1.2\textwidth}{Table 4. Number (frequency) of $P_s$ and $P_l$ under 320 $\times$ 800 resolution.}}}
        \vspace{-4pt}
        \setlength{\tabcolsep}{1.2mm}
        \renewcommand{\arraystretch}{1.2}{
        \scalebox{0.67}{
        \begin{tabular}{c |c c}
            \hline
            Methods & $P_s$ & $P_l$ \\
            \hline
            Ours-fast & 2654 (44.1\%) & 3365 (55.9\%) \\
            Ours-faster & 2565 (42.6\%) & 3454 (57.4\%) \\
            \hline
        \end{tabular}
        \label{DistributionTab}
        }}
    \end{minipage}
    \end{figure}

\vspace{-12pt}
\paragraph{Analysis of patch size distribution.} We analyze the distribution of dynamically selected patch sizes to better understand the behavior of SPSS in practical scenarios. Fig. \ref{DistributionFig} illustrates the temporal evolution of selected patch sizes together with the average object depth in a representative nuScenes scene (ID: \texttt{9f1f69646d644e35be4fe0122a8b91ef}), where a clear positive correlation can be observed. Specifically, as the average object depth increases, SPSS tends to select larger patch sizes, indicating that distant-dominant scenes are processed with coarser patches to reduce redundant background computation.
Conversely, scenes with closer objects are more likely to be assigned smaller patch sizes to preserve fine-grained semantic details. To further quantify this behavior, Tab. \ref{DistributionTab} reports the frequency of selected small ($P_s$) and large ($P_l$) patch sizes on the nuScenes validation set under $320 \times 800$ resolution.
Both SEPatch3D-\textit{fast} and SEPatch3D-\textit{faster} variants exhibit a diverse usage of $P_s$ and $P_l$, rather than collapsing to a single patch size. This confirms that SPSS performs effective dynamic selection instead of relying on a fixed configuration.
Overall, these results demonstrate that the proposed spatiotemporal-aware patch size selection adapts to scene-level depth variations in a stable manner, validating its practical effectiveness and robustness across diverse driving scenarios.

\vspace{-12pt}
\paragraph{Impact of the adaptive patch selection strategy in the IPS module.}
As shown in Tab. \ref{mean-entropy-based}, fixed-ratio Top-\textit{K} selection performs worse than our adaptive strategy in both accuracy and latency. We attribute this to its inability to accommodate diverse scene conditions. When the keep ratio is too small, many patches that require refinement fail to be selected, leading to missing detail enhancement. Conversely, when the keep ratio is large, redundant patches are included, introducing unnecessary computation and slowing inference. These results demonstrate the necessity and effectiveness of our adaptive patch selection strategy.

    \begin{table}[ht]
        \centering
        \caption{Ablations of adaptive patch selection strategy in the IPS module.}
        \setlength{\tabcolsep}{0.5mm}{
        \scalebox{0.82}{
        \begin{tabular}{c|c|c c|c} 
        \hline
        Strategies & Keep Ratio & NDS (\%) $\uparrow$ & mAP (\%) $\uparrow$ & Infe. Time (ms) $\downarrow$  \\
        \hline
        - (Baseline) & - & 61.2 & 52.1 & 317.0 \\
        \hline
        \multirow{3}{1.0cm}{Top-\textit{K}} & 30\% & 59.6 & 51.0 & 198.4 \\
        & 50\% & 59.8 & 51.1 & 202.7 \\
        & 70\% & 59.9 & 51.3 & 207.6 \\
        \hline
        Adaptive (Ours) & - & 60.3 & 51.6 & 194.3 \\
        \hline
        \end{tabular}
            }
        }
        \label{mean-entropy-based}
    \end{table}

\vspace{-12pt}
\paragraph{Impact of the IPS module placement.}
As shown in Tab. \ref{IPS-placement}, placing the IPS module after the fine patches yields the best trade-off between accuracy and efficiency. Applying IPS only after coarse patches slightly improves accuracy but introduces additional computation, resulting in 14.9\,ms increase in inference time. This is because selection performed on coarse patches must later be mapped back to the fine patches for enhancement, which inevitably brings in irrelevant fine patches that were not truly informative, thereby adding unnecessary computation. Applying IPS after both fine and coarse patches further increases redundancy, offering negligible accuracy gains while significantly slowing down inference. These results indicate that informative patch selection is most effective when performed on fine patch features, where informative details are better preserved and redundant patches can be avoided. Hence, we adopt placing IPS after the fine patches as the default design.

    \begin{table}[ht]
        \centering
        \caption{Ablations of the IPS module placement.}
        \setlength{\tabcolsep}{0.7mm}{
        \scalebox{0.75}{
        \begin{tabular}{c|c c|c} 
        \hline
        Position & NDS (\%) $\uparrow$ & mAP (\%) $\uparrow$ & Infe. Time (ms) $\downarrow$  \\
        \hline
        After fine patches only & 60.3 & 51.6 & 194.3 \\
        After coarse patches only & 60.2 & 51.7 & 209.2 \\
        After both fine and coarse patches & 60.4 & 51.7 & 213.8 \\
        \hline
        \end{tabular}
            }
        }
        \label{IPS-placement}
    \end{table}

\section{More Visualization Results} \label{More visualization results}
\begin{figure}[ht]
        \centering
        \captionsetup{skip=2pt}
        \includegraphics[width=1.0\linewidth]{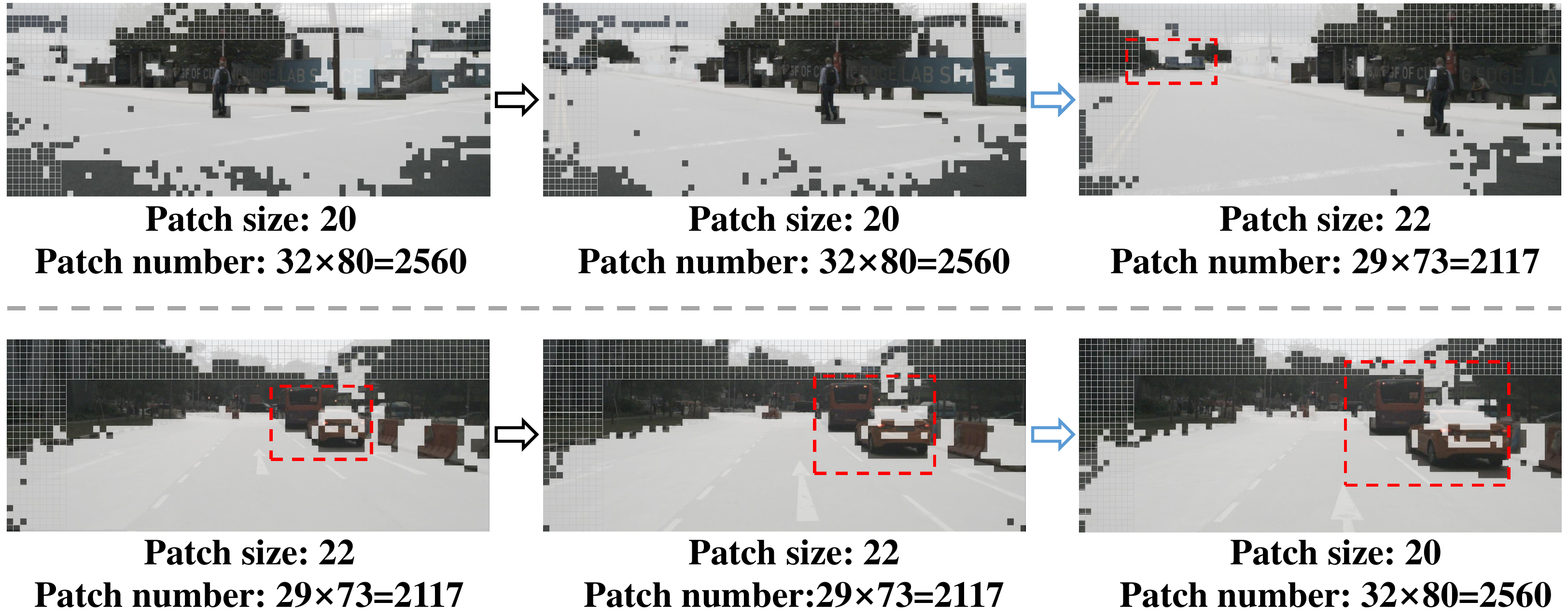}
        \caption{Visualization of the patch size selection process in the SPSS module. The blue arrows mark the frames where the patch size is adjusted. The red boxes indicate the regions exhibiting noticeable changes.}
        \label{SPSS_vis}
    \end{figure}

In Fig. \ref{SPSS_vis}, we show the partial patch grids overlaid on the upper and left regions of the images. In the first row, the patch size increases from 20 to 22 when the ego-car turns into a new road segment. This adjustment occurs because new and farther objects (marked by the red box) appear in the new road segment. In contrast, the second row depicts the patch size decreasing from 22 to 20 as the objects gradually approach and occupy a larger region of the image. The visualizations indicate that SPSS dynamically adjusts patch sizes according to the spatiotemporal changes of objects, confirming its effectiveness.

\begin{figure}[ht]
        \centering
        \captionsetup{skip=2pt}
        \includegraphics[width=1.0\linewidth]{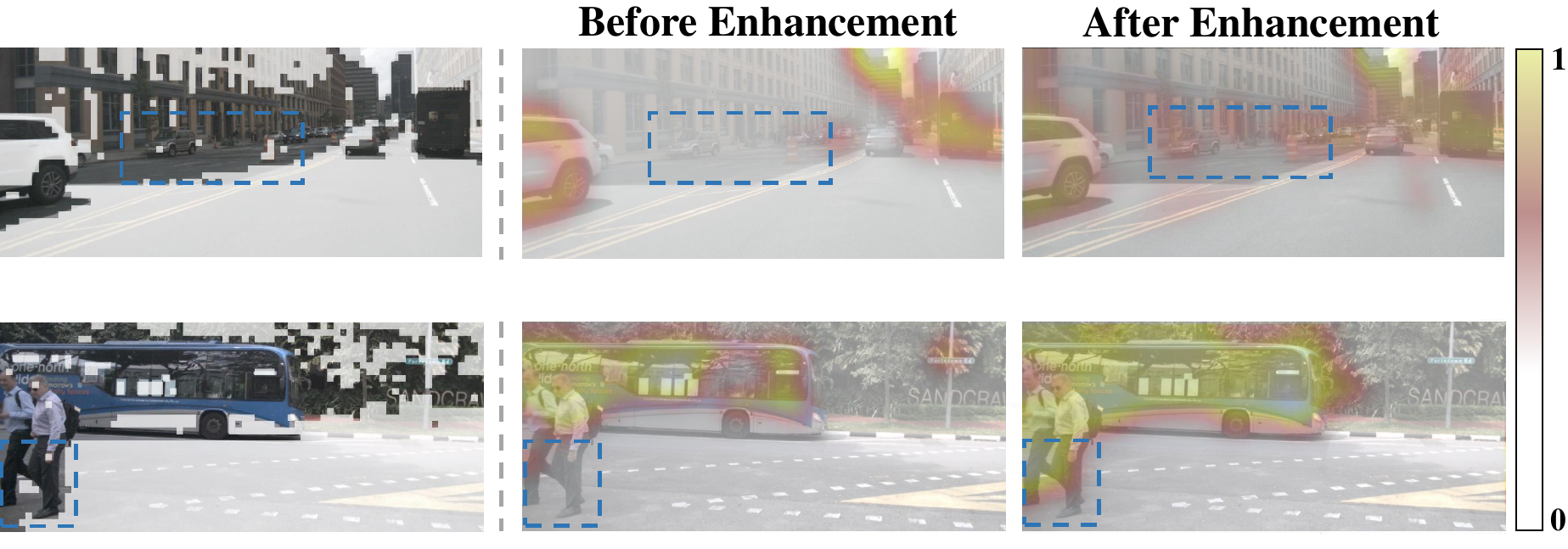}
        \caption{More visualizations of informative regions and feature responses.}
        \label{more_vis}
    \end{figure}

In Fig. \ref{more_vis}, we provide additional visualizations of the selected tokens and feature responses, further extending the analyses presented in the main manuscript to confirm that our IPS module consistently focuses on informative regions and our CGFE module further enhances the feature representations within these selected patch regions.

\begin{figure}[ht]
        \centering
        \captionsetup{skip=2pt}
        \includegraphics[width=1.0\linewidth]{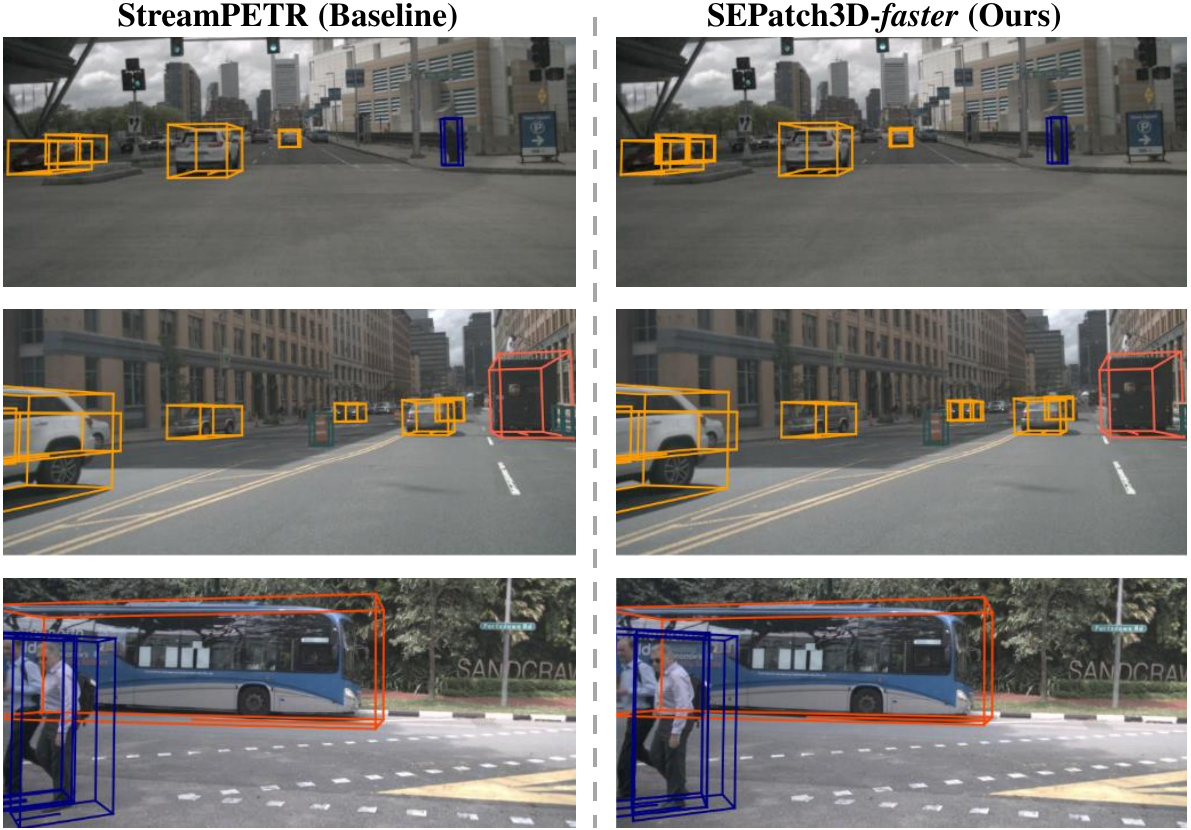}
        \caption{Visualization of detection results between StreamPETR (baseline) and our SEPatch3D-\textit{faster}. Bounding box colors indicate different object categories.}
        \label{results_vis}
    \end{figure}

In Fig. \ref{results_vis}, we compare the detection results between the baseline and our SEPatch3D-\textit{faster}. The results indicate that our faster variant achieves comparable detection quality without notable missed detections, demonstrating the effectiveness of our approach in maintaining accuracy while improving efficiency.

\section{Limitation and Future Work} \label{limitation}
While SEPatch3D effectively improves efficiency by adaptively adjusting patch sizes, the current patch size selection is still based on the pre-defined heuristic priors. In complex scenes with highly dynamic layouts, such heuristics may not always yield the optimal granularity. In future work, we plan to explore learnable patch size selection strategies that allow the network to automatically determine patch sizes in a data-driven manner. Moreover, we plan to further accelerate ViT-based multi-view 3D detectors by incorporating quantization or even binarization techniques \cite{gao2025bhvit}, enabling real-time perception while maintaining competitive accuracy. Beyond 3D detection, we also plan to extend our SEPatch3D to other query-based spatiotemporal scene understanding tasks, such as 3D occupancy prediction \cite{li2026ashsr} and world model \cite{zheng2025world4drive}.

{
    \small
    \bibliographystyle{ieeenat_fullname}
    \bibliography{main}
}


\end{document}